\documentstyle[12pt]{article}

\newcommand{\citename}[1]{{#1}}

\begin{document}

\title{
{\bf GRAMPAL: A Morphological Processor for Spanish implemented in
Prolog}\footnote{This work has been partially supported by the Spanish {\em
Plan Nacional de I+D}, under the Research Project {\em An Architecture for
para Natural Language Interfaces with User Modeling\/} (TIC91-0217C02-01).}
}

\author{\normalsize
\parbox[t]{70mm}{
\begin{center}
{\bf Antonio Moreno} \\
{\em Dep. de Ling\"{u}\'{\i}stica} \\
{\em Universidad Aut\'{o}noma de Madrid} \\
{\em Cantoblanco, Madrid, SPAIN} \\
{\em Phone: +34 1  397.41.09} \\
{\em Fax: +34 1  397.39.30} \\
{\tt sandoval@ccuam3.sdi.uam.es} \\
\end{center}
}\parbox[t]{75mm}{
\begin{center}
{\bf Jos\'{e} M. Go\~{n}i} \\
{\em Dep. Matem\'{a}tica Aplicada a las TT.II.} \\
{\em Universidad Polit\'{e}cnica de Madrid} \\
{\em Ciudad Universitaria, Madrid, SPAIN} \\
{\em Phone: +34 1  336.72.87} \\
{\em Fax: +34 1  336.72.89} \\
{\tt jmg@mat.upm.es} \\
\end{center}
}
}

\date{}
\maketitle
\thispagestyle{empty}

\begin{abstract}

A model for the full treatment of Spanish inflection for verbs, nouns and
adjectives is presented. This model is based on feature unification and it
relies upon a lexicon of allomorphs both for stems and morphemes.  Word
forms are built by the concatenation of allomorphs by means of special
contextual features. We make use of standard Definite Clause Grammars (DCG)
included in most Prolog implementations, instead of the typical
finite-state approach. This allows us to take advantage of the
declarativity and bidirectionality of Logic Programming for NLP.

The most salient feature of this approach is simplicity: A really
straightforward rule and lexical components. We have developed a very simple
model for complex phenomena.

Declarativity, bidirectionality, consistency and completeness of the model
are discussed: all and only correct word forms are analysed or generated,
even alternative ones and gaps in paradigms are preserved.  A Prolog
implementation has been developed for both analysis and generation of
Spanish word forms.  It consists of only six DCG rules, because our {\em
lexicalist\/} approach --i.e. most information is in the
dictionary. Although it is quite efficient, the current implementation
could be improved for analysis by using the non logical features of Prolog,
especially in word segmentation and dictionary access.

\end{abstract}


\section{Introduction}

\hspace{3.3mm}
The successful treatment of morphological phenomena in some languages by
means of finite state automata appears to have led to the idea that this
model is the most efficient and universal way to deal with morphology
computationally.  Although there exist good finite-state processors for
Spanish --like \cite{Marti1986}, \cite{Meya1986} or
\cite{Tzoukermann-Liberman1990}-- we think that some phenomena can be
handled more elegantly using a context-free approach, particularly if the
morphological component is to be included as a part of a syntax
grammar. Our model has been implemented in standard DCG using a logic
programming approach instead of a plain finite-state one.

It is well-known that the so-called non-concatenative processes are the
most difficult single problem that morphological processors must deal with.
Experience has shown that it is not easy for any approach.
Unification-based morphology uses suppletion (i.e. alternative allomorphs
for a lemma) and feature description as a general mechanism for handling
those processes.  Two-level morphology uses instead rules that match
lexical representations (lemmas) with surface representations (actual
spelling forms).  The latter has been been claimed to be more elegant, but
it is obvious that often the two-level model contains many rules needed for
a very few cases.

The pure two-level/finite-state automata model is not very adequate for
treating certain non-concatenative processes, and in such cases one is
required to depart from this approach, for example by adding an extension
in which two-level rules are retained under the control of feature
structures \cite{Trost1990}.  Moreover, every language has irregularities
that can only be treated as suppletive forms, e.g. {\tt soy} ({\em I
am\/}), {\tt era} or {\tt fui} ({\em I was\/}).  Since suppletion is needed
anyway, and since it is a much simpler approach than rules, we consider
that the ``elegance'' objection is not well-founded\footnote{See the next
section for further discussion of the adequacy of the two-level model for
Spanish, including defective forms (i.e. null forms in the conjugation) and
alternative correct forms.}.

On the other hand, our goal is to generate and recognize all (and only)
well-formed inflected forms, and thus we do not accept ``{\em missing
forms\/}'' for defective verbs (see below), but do accept duplicate but
correct forms.

\section{Major issues in Spanish computational morphology}
\hspace{3.3mm}
Spanish morphology is not a trivial subject.  As an inflectional
language, Spanish shows a great variety of morphological processes,
particularly non-concatenative ones.  We will try to summarize the
outstanding problems which any morphological processor of Spanish has to
deal with:

\begin{enumerate}

\item A highly complex verb paradigm. For simple tenses, we consider 53
inflected forms (see Table~\ref{table1}), excluding the archaic
Future Subjunctive, but including the duplicate Imperfect Past Subjunctive
(6 forms). If we add the 45 possible forms for compound tenses, then 98
inflected forms are possible for each verb.

\item  The frequent irregularity of both verb stems and endings.  Very common
verbs, such as {\tt tener} ({\em to have\/}), {\tt poner} ({\em to put\/}),
{\tt poder} ({\em to be able to\/}), {\tt hacer} ({\em to do}), etc., have up
to 7 different stems: {\tt hac-er}, {\tt hag-o}, {\tt hic-e}, {\tt ha-r\'{e}},
{\tt hiz-o}, {\tt haz}, {\tt hech-o}. This example shows internal vowel
modification triggered by different morphemes having the same external form:
{\tt hag-o}; {\tt hiz-o}, {\tt hech-o} (The first {\tt /-o/} is first person
singular present indicative morpheme; the second {\tt /-o/} is third singular
preterit indicative morpheme; and the third {\tt /-o/} is past participle
morpheme -- an irregular one, by the way).  As well as
these non-concatenative
processes, there exist other, very common, kinds of internal variation,
as illustrated by the following example.

\begin{center}
\begin{tabular}{llccc}
$[e] \rightarrow [ie]$: & \makebox[5mm]{} & {\tt quer-er} ({\em to want\/}) &
 $\rightarrow$ & {\tt quier-o}  ({\em I want\/})
\end{tabular}
\end{center}

2,300 out of 7,600 verbs in our dictionary are classified as irregular, and
5,300 as regular --i.e. only one stem for all the forms as in {\tt am-ar}
--- {\tt am-o}, etc. ({\em to love\/}).

\item Gaps in some verb paradigms. In the so-called {\em defective verbs}
some forms are missing or simply not used. For instance, meteorological verbs
such as {\tt llover}, {\tt nevar} ({\em to rain}, {\em to snow\/}), etc. are
conjugated only in third person singular. Other ones are more peculiar, like
{\tt abolir} ({\em to abolish\/}) that lacks first, second and third singular
and third plural present indicative forms, all present subjunctive forms, and
the second singular imperative form. In other verbs, the compound tenses are
excluded from the paradigm, like in {\tt soler} ({\em to do usually\/}).

\item Duplicate past participles:  a number of verbs have two
alternative forms, both correct, like {\tt impreso}, {\tt imprimido}
({\em printed}).  In such cases, the analysis has to treat both.

\item There exist some highly irregular verbs that can be
handled only by including many of their forms directly in the lexicon (like
{\tt ir} ({\em to go\/}), {\tt ser} ({\em to be\/}), etc).

\item Nominal inflection can be of two major types: with
grammatical gender (i.e. concatenating the gender morpheme to the stem) and
with inherent gender (i.e. without gender morphemes). Most pronouns and
quantifiers belong to the first class, but nouns and adjectives can be in
any of the two classes, with a different distribution: 4\% of the nouns
have grammatical gender and 92\% have inherent gender, while 70\% of the
adjectives are in the first group.  Some nouns and adjectives present
alternative correct forms for plural --e.g. for {\tt bamb\'{u}} ({\em
bamboo\/}), {\tt bamb\'{u}-s} and, {\tt bamb\'{u}-es}.

\item There is a small group (3\%) of invariant nouns with the same form for
singular and plural, e.g. {\tt crisis}. On the other hand, 30\% of the
adjectives present the same form for masculine and feminine, e.g. {\tt azul}
({\em blue\/}).  There exist also {\em singularia tantum\/}, where only the
singular form is used, like {\tt estr\'{e}s} ({\em stress\/}); and {\em
pluralia tantum\/}, where only the plural form is allowed, e.g. {\tt
matem\'{a}ticas}, ({\em mathematics\/}).

\item In contrast with verb morphology, nominal processes do not produce
internal change in the stem caused by the addition of a gender or plural
suffix, although there can be many allomorphs produced by spelling changes:
{\tt luz}, {\tt luc-es} ({\em light, lights\/}).

\end{enumerate}

For a detailed description of all verb and nominal phenomena, including a
classification into paradigmatic models, see \cite{Mor91}.

All these phenomena suggest that there is no such a universal model
(e.g. two-level, unification, or others) for (surface) morphology.
Instead, we have approaches more suited for some processes than others.
The computational morphologist must decide which is more appropriate for a
particular language. We support the idea that unification and feature-based
morphology is more adequate for languages, such as Spanish and other Latin
languages, that have alternative stems triggered by specific suffixes,
missing forms in the paradigm, and duplicate correct forms.

\section{The model}
\hspace{3.3mm}
It is well known that morphological processes are divided into two types:
processes related to the phonological and/or graphic form (morpho-graphemics),
and processes related to the combination of morphemes (morpho-syntax). Each
model treats these facts from its particular perspective. Two-level morphology
uses phonological rules and continuation classes (in the lexical component).
Mixed systems such as \cite{Bear1986} or \cite{RitchieEtAl1987} have different
sets of rules.

As we stated before, our model relies on a context-free feature-based grammar,
that is particularly well suited for the morpho-syntactic processes.  For
morpho-graphemics, our model depends on the storage --or computation-- of all
the possible allomorphs both for stems and endings. This feature permits that
both analysis and synthesis be limited to morpheme concatenation, as the
general and unique mechanism. This simplifies dramatically the rule component.

We present some examples of dictionary entries: two verbal ending entries
(allomorphs) for the past participle morphemes and two allomorph stems for
{\em imprimir}, compatible with those endings.

\begin{quote}{\begin{verbatim}
vm(no,part,nofin,[2,3],99,reg) --> [ido].
vm(no,part,nofin,[2,3],99,part1) --> [o].

vl(imprimir,v,3,[100],[reg]) --> [imprim].
vl(imprimir,v,3,[99],[part1]) --> [impres].
\end{verbatim}
}\end{quote}

Where {\tt vm} and {\tt vl} stands for the values of the ``morphological
category'' that we are using to drive the DCG rule invocation. All the
dictionary entries are coded with a predicate that corresponds to its
morphological category. The full inventory of such categories follows:

\begin{description}
\item[w] For complete inflected {\bf w}ord forms.
\item[wl] For words (nouns and adjectives) that can accept a number
morpheme.
\item[vl] For {\bf v}erb {\bf l}exemes (stems).
\item[nl] For {\bf n}ominal --nouns and adjectives-- {\bf l}exemes.
\item[vm] For {\bf v}erb {\bf m}orphemes.
\item[ng] For {\bf n}ominal {\bf g}ender morphemes.
\item[nn] For {\bf n}ominal {\bf n}umber morphemes.

\end{description}

For reference, and to check the meaning of the examples, a short
self-description of the arguments of those predicates follows:

\begin{quote}{\begin{verbatim}
w(Lemma, Category, Pers_Num, Tense_Mood).
w(Lemma, Category, Gender, Number).
wl(Lemma, Category, Number_Type_List, Gender, Number).

vl(Lemma, Category, Conjugation, Stem_Type_List,
   Suffix_Type_List).
vm(Pers_Num, Tense_Mood, Finiteness, Conjugation_List,
   Stem_Type, Suffix_Type).

nl(Lemma, Category, Gender_Type_List, Number_Type_List,
   Gender, Number).
ng(Gender_Type, Gender, Number).
nn(Number_Type, Number).

\end{verbatim}
}\end{quote}

We have introduced some contextual atomic features that impose restrictions
on the concatenation of morphemes through standard unification rules. Such
features are never percolated up to the parent node of a rule. Multi-valued
atomic features are permitted in the unification mechanism, being
interpreted as a disjunction of atomic values. We represent this
disjunction as a Prolog list. Disjunction of values is used only for
contextual features ({\em stem\_type, suffix\_type, conjugation,
gender\_type} and {\em number\_type}) just to improve storage efficiency,
since this device is actually not needed if different entries were encoded
in the lexicon.

In the conjugation table (Table~\ref{table1}), the {\em stem\_type} values
of the grammatical features person-number and tense-mood are displayed in
boldface. For example, {\bf sing\_1} means first person, singular number;
while {\bf pres\_ind} means present tense, indicative mood.

\begin{quote}{
\begin{table}[htb]
\begin{center}
\begin{tabular}{|c||c|c|c|c|c|c|c|}\hline
        & {\bf sing\_1} & {\bf sing\_2} & {\bf sing\_3} &
          {\bf plu\_1} & {\bf plu\_2} & {\bf plu\_3} & --- \\ \hline \hline
{\bf pres\_ind}     & 11 & 12 & 13 & 14 & 15 & 16 &     \\ \hline
{\bf impf\_ind}     & 21 & 22 & 23 & 24 & 25 & 26 &     \\ \hline
{\bf indf\_ind}     & 31 & 32 & 33 & 34 & 35 & 36 &     \\ \hline
{\bf fut\_ind}      & 41 & 42 & 43 & 44 & 45 & 46 &     \\ \hline
{\bf pres\_subj}    & 51 & 52 & 53 & 54 & 55 & 56 &     \\ \hline
{\bf impf\_subj}    & 61 & 62 & 63 & 64 & 65 & 66 &     \\ \hline
{\bf cond}          & 71 & 72 & 73 & 74 & 75 & 76 &     \\ \hline
{\bf imper}         &    & 82 & 83 &    & 85 & 86 &     \\ \hline
{\bf inf}           &    &    &    &    &    &    &  00 \\ \hline
{\bf ger}           &    &    &    &    &    &    &  90 \\ \hline
{\bf part}          &    &    &    &    &    &    &  99 \\ \hline
\end{tabular}
\end{center}
\caption{Conjugation table.}\label{table1}\end{table}
}\end{quote}

Each of the 49 inflected forms\footnote{The codes 83 and 86 stand for the
{\em courtesy} imperative: {\tt imprima usted}, {\tt impriman
ustedes}. These word forms are the same as the 53 and 56 ones.} is
represented by a numeric code\footnote{Actually, this number encodes a
particular combination of person, number, tense and mood features.}, and
the additional value {\tt 100} is used as a shorthand for the disjunction
of all of them (used for regular verbs; see the entry for {\tt imprim}
above). The contextual feature {\em stem\_type\/} ({\bf stt}) is used to
identify the verb stem and ending corresponding to each form, and the
contextual feature {\em suffix\_type\/} ({\bf sut}) distinguishes among
several allomorphs of the inflectional morpheme by means of a set of
values:

\begin{quote}{
\begin{center}
\begin{tabular}{llll}
{\tt reg} & {\tt pres} & {\tt pret1}  & {\tt pret2} \\
{\tt fut\_cond} & {\tt imp\_subj} & {\tt imper} & {\tt infin} \\
{\tt ger} & {\tt part1} & {\tt part2} & \\
\end{tabular}
\end{center}
}\end{quote}

Since this value set is much smaller than the {\em stem\_type set}, we have
chosen an alphabetic code. With the combination of both features, and the
addition of a third feature {\tt conj} (conjugation), we can state
unequivocally which is the correct sequence of stem and ending for each case
(see examples above, where {\tt imprim} only matches {\tt ido} for all
features, and {\tt impres} matches {\tt o}, thus preventing ill-formed
concatenations --for these morphemes-- such as {\tt imprim-o} or {\tt
impres-ido}).

In the same fashion, we have two special contextual features for the
nominal inflection, {\tt nut} (number\_type) and {\tt get} (gender\_type),
to identify the various allomorphs for the plural and gender morphemes, and
associate them with the proper nominal stems. The following examples show
those contextual features both in nominal morphemes and in nominal lexeme
entries:

\begin{quote}{\begin{verbatim}
/*     NOUN AND ADJECTIVE MORPHEMES     */
ng(mas1,masc,sing) --> [o].
ng(mas2,masc,sing) --> [e].
ng(fem,fem,sing)   --> [a].

nn(plu1,plu) --> [s].
nn(plu2,plu) --> [es].

/*     SOME ENTRIES FOR NOUNS     */
nl(presidente, n, [mas2,fem], [plu1], _, _) --> [president].

wl(doctor, n, [plu2], masc, sing) --> [doctor].
nl(doctor, n, [fem], [no], masc, sing) --> [doctor].

wl(bambu1,  n, [plu1,  plu2],  masc,  sing) --> [bambu1].

\end{verbatim}
}\end{quote}

These entries allow the analysis/generation of the word forms {\tt
presidente, presidenta, presidentes} and {\tt presidentas} for the lemma
{\tt presidente}; {\tt doctor, doctora, doctores} and {\tt doctoras} for
{\tt doctor}; and {\tt bamb\'{u}, bamb\'{u}s} and {\tt bamb\'{u}es} for {\tt
bamb\'{u}}.

The grammatical features ({\em category, lemma, tense, mood, person,
number\/} and {\em gender\/} are the only features that are delivered to
the {\bf w} node, and from this level can be used by a syntactic DCG
grammar.

A unification-based system relies very much on the lexical side. It is
needed a robust and large dictionary, properly coded. Additionally, our
model depends on the accessibility of all possible allomorphs, so their
storage is also necessary. Fortunately, there is no need for typing all of
them by hand, since this would be an impractical, time consuming and
error-prone task.  Morpho-graphemics for Spanish is quite regular and we
have formalized and implemented the automatic computation of the allomorphs
of any verb from the infinitive form.

The formalized description of the morphological phenomena of Spanish was
presented in \cite{Mor91}, where some interesting and well founded
linguistic generalizations are made: Paradigms\footnote{These are not the
traditional ones, since they capture the problems arising in written
language, such as diacritical marks, different surface letters for the same
phoneme, etc.} for verbs are described to capture regularities in the
inflectional behaviour of the Spanish verbs, and the same is done with
nouns.  All the {\em lemmas\/} belonging to a particular paradigmatic model
not only share most of contextual and grammatical features but also have
the same allomorph number and distribution. For instance, our model 11
has three allomorph stems, and their distribution is as follows:

\begin{quote}{\begin{verbatim}
vl(salir, 3, [0,12,13,14,15,16,21,22,23,24,25,26,
              31,32,33,34,35,36,61,62,63,64,65,66,
              82,85,90,99],[reg])                   --> [sal].
vl(salir,3,[11,51,52,53,54,55,56,83,86],[reg])      --> [salg].
vl(salir,3,[41,42,43,44,45,46,71,72,73,74,75,76],
           [fut_cond])                              --> [sald].
\end{verbatim}}\end{quote}

In \cite{Goni1994} regular-expression based rules are devised to
compute automatically these allomorphs, capturing morpho-graphemic
generalizations in the paradigmatic models.

\section{The grammar}
\hspace{3.3mm}
The rule component of the model is quite small, because most of the
information is in the lexicon. In particular, inflected verb forms are
analysed or generated by two rules. Actually, only one rule is needed, but as
we used the value {\tt 100} for the {\tt stt} feature for regular verbs
instead of a disjunction of all the possible {\tt stt} values, we split the
rule in two:

\begin{quote}{\begin{verbatim}
/**/
/*           VERB INFLECTION RULES          */
/**/
w(Lex, Cat, PerNum, TensMood) -->
            vl(Lex, Cat, Conj, SttL, SutL),
            vm(PerNum, TensMood, _,  ConjL, Stt, Sut),
        {
                member(Conj,ConjL),
                member(Stt, SttL),
                member(Sut,SutL)
        }.

w(Lex, Cat, PerNum, TensMood) -->
            vl(Lex, Cat, Conj, [100], SutL),
            vm(PerNum, TensMood, _, ConjL, _, Sut),
        {
                member(Conj,ConjL),
                member(Sut,SutL)
        }.

\end{verbatim}
}\end{quote}

Nominal inflection is a bit more complicated, because of the combination of
two inflectional morphemes (gender and number) in some cases.  Our model
needs the 4 rules shown to handle this. The first one is for singular
words, when the stem has to be concatenated to a gender suffix ({\tt
ni\~{n}-o, ni\~{n}-a}); the second is for plural words, where an additional
number suffix is added ({\tt ni\~{n}o-s}) ; the third builds plurals from an
allomorph stem and a plural morpheme ({\tt le\'{o}n / leon-es}); and the
fourth rule validates as words the singular forms ({\bf wl}) obtained from
the first rule without further concatenation:

\begin{quote}{\begin{verbatim}
/**/
/*     NOUN AND ADJECTIVE INFLECTION RULES   */
/**/

wl(Lex, Cat, [plu1], Gen, Num)
        --> nl(Lex, Cat, GetL, _, _, _), ng(Get, Gen, Num),
        {
                member(Get, GetL)
        }.

w(Lex, Cat, Gen, Num) -->
        wl(Lex, Cat, NutL, Gen, _), nn(Nut, Num),
        {
                member(Nut, NutL)
        }.

w(Lex, Cat, Gen, Num) -->
        nl(Lex, Cat, _, NutL, Gen, _), nn(Nut, Num),
        {
                Nut = plu2,
                member(Nut, NutL)
        }.

w(Lex, Cat, Gen, Num) -->
        wl(Lex, Cat, _, Gen, Num).

\end{verbatim}
}\end{quote}

The predicate {\tt member} included in the procedural part of the DCG rule
implements disjunction in atomic contextual features, although it could
have been eliminated with a different encoding of the lexical entries.

\section{The Processor}
\hspace{3.3mm}
The grammar rules are stated using the DCG formalism included in most
Prolog implementations, thus we have used the DCG interpreter both for
parsing and generating word forms. Since the interpreter is supplied with
morphemes included in the dictionary for its proper operation, a segmenter
has to be included to provide the parser with candidate word
segmentations. This is achieved by means of a non-deterministic predicate
that finds all the possible segmentations of a word. This is one of the
efficiency drawbacks of the current implementation of GRAMPAL.

To avoid such inefficiency the system could be augmented with a {\em letter
trie\/} index --or {\em trie\/}-- \cite{AhoEtAl1983} to the lexical
entries. With this device, segmentation will be no longer
non-deterministically blind and the search would be efficiently guided.
Generation does not have those efficiency problems, and the system is
bidirectional without any change in the rules.

\section{Conclusions}
\hspace{3.3mm}
A Prolog prototype, GRAMPAL, was developed to intensively test the model,
both as analyser and as generator. This processor implemented in Prolog has
shown that logic programming can be used successfully to handle the Spanish
inflectional morphology. We have also implemented a C version of GRAMPAL,
but it needs separate components for analysis and generation, due to the
lack of reversibility that Logic Programming has provided us with.

The model presented is based on two basic principles:

\begin{itemize}

\item Empirical rigour: all and only correct forms are analysed and
generated, whether regular or not; gaps in verb paradigms are observed;
suppletive forms are considered valid, and so on. It is important to
stress that GRAMPAL does not overgenerate or overanalyse.

\item Simplicity and generalization: GRAMPAL employs a really
straightforward rule component, that captures the logical generalization of
the combination of a stem and an ending to form a inflected
word. ``Standard scientific considerations such as simplicity and
generality apply to grammars in much the same way as they do to any other
theories about natural phenomena. Other things being equal, a grammar with
seven rules is to be preferred to one with 93 rules''  \cite{Gazdar89}.

\end{itemize}

The current dictionary has a considerable size: 43,000 {\em lemma\/}
entries\footnote{In these figures are neither included closed categories,
nor allomorphs for verb and nominal morphemes.}, including 24,400 nouns,
7,600 verbs, and 11,000 adjectives.  The model could be used for derivative
morphology and compounds as well, but this has not been done yet, since
further linguistic analysis must be done to specify the features needed to
permit derivatives and compounds.

\end{document}